**Enhancing Systematic Reviews with Large Language Models: Using GPT-4 and Kimi**

Dandan Chen Kaptur, Ph.D., Pearson
Yue Huang, Ph.D., Measurement Incoporated
Xuejun Ryan Ji, Ph.D., British Columbia College of Nurses and Midwives
Yanhui Guo, Ph.D., University of Illinois Springfield
Bradley Kaptur, M.D., HSHS Saint John's Hospital

Paper presented at the National Council on Measurement in Education (NCME) Conference, Denver, Colorado, in April 2025.

**Enhancing Systematic Reviews with Large Language Models: Using GPT-4 and Kimi**

**Abstract**

This research delved into GPT-4 and Kimi, two Large Language Models (LLMs), for systematic reviews. We evaluated their performance by comparing LLM-generated codes with human-generated codes from a peer-reviewed systematic review on assessment. Our findings suggested that LLMs' performance fluctuates by data volume and question complexity for systematic reviews.

*Word count*: 785

## Introduction

Despite the growing use of Large Language Models (LLMs) in academic research, questions about their accuracy and reliability persist. This study compares GPT-4, a widely used LLM, and Kimi, less studied particularly in Western academia. We conducted a comparative analysis of GPT-4 and Kimi, to explore nuances in their performance for different situations in the context of text coding for systematic reviews.

### Methods

We fed 32 articles covered by a systematic review on technology-based assessments (Chen et al., 2023) into LLMs using specific yes-or-no questions from the original review. To account for variation in LLM performance due to data volume, we ran the coding process using three batch sizes: single, four, and eight articles. To address variation due to question complexity, we used 11 questions from the review. This coding process involves 33 conditions (3 batch sizes × 11 questions), resulting in 1,056 values (32 articles × 33 conditions) for coding, where the LLMs responded with either 1 ("Yes") or 0 ("No") for each question.

The 11 questions, shown in Table 1, were the original ones from the review, not modified to allow comparison between LLM-generated codes and the human-generated codes from the review. The articles in PDF format, were uploaded as attachments and coded based on these questions submitted together in one prompt.

We evaluated LLM performance using standard metrics: accuracy, precision, recall, and F1 scores. Details are available in Appendix A. The original human-generated codes were treated as the true classifications, and the LLM-generated codes served as predictions. To minimize the effect of random chance, we replicated the LLM coding process four times, with each coder using a different computer at a different time. We ended up with one set of human-generated codes and four sets of LLM-generated codes for each condition.

Inter-coder reliability was measured using the Intraclass Correlation Coefficient (ICC) and Kappa statistics to assess consistency among the five sets of 0-1 codes across all conditions.

**Results**

Table 2 summarizes the overall performance of GPT-4 and Kimi across all conditions. GPT-4 slightly outperforms Kimi in accuracy (0.660 vs. 0.633), recall (0.791 vs. 0.712), and F1 score (0.785 vs. 0.748). These results indicate that GPT-4 was more effective in correctly identifying relevant content and maintains a better balance between precision and recall. However, Kimi demonstrated slightly higher precision (0.788 vs. 0.779), suggesting it is better at identifying true positives.

Table 3 details performance by batch size. When coding one article at a time, GPT-4 and Kimi had similar accuracy (0.738 vs. 0.744) and F1 scores (0.803 vs. 0.799). Kimi outperformed

GPT-4 in precision (0.811 vs. 0.780), but GPT-4 showed better recall (0.827 vs. 0.787). As the batch size increased, GPT-4 maintained higher accuracy, recall, and F1 scores, while Kimi's precision advantage diminished, disappearing when coding eight articles (0.766 vs. 0.782).

Table 4 assesses inter-coder reliability. Both models show strong agreement when coding one article, with Kimi achieving higher ICC (0.845 vs. 0.815) and Kappa (0.521 vs. 0.468). For larger batches, GPT-4's consistency remained high, while Kimi's reliability declined below 0.75 when the batch size was eight.

Item-specific performance, presented in Table 5, reveals that GPT-4 and Kimi performed well on inclusion criteria items (Items 1-5), with Kimi outperforming GPT-4 in some cases. For synthesis items (Items 6-11), both models showed lower accuracy and recall. However, Kimi generally performed better than GPT-4 in precision, as shown in Figure 1 when coding one article at a time.

## Discussion

LLMs like GPT-4 and Kimi showed promise in assisting systematic reviews, particularly by reducing workload during initial coding phases and pre-screening human-coded results for potential errors. This study compared GPT-4 and Kimi's performance in coding 32 articles using 11 fixed yes-or-no questions from a systematic review on technology-based assessments.

Our findings show that GPT-4 performed better overall and remained stable as batch size increased, while Kimi's performance declined with larger batches. Kimi reached its optimal performance when the batch size was one, showing it was more effective at identifying true positives in most cases.

However, neither model achieved a sufficient threshold of performance for unsupervised use in systematic reviews. Both GPT-4 and Kimi struggled with consistent performance across various questions, indicating that further research is needed to understand how item features affect coding outcomes.

Future studies should explore prompt engineering to improve the quality of questions we feed into LLMs and see how it can enhance coding accuracy. They could involve other models like Claude, which has been gaining attention for its capabilities. These explorations could shed light on how we could optimize LLM performance in systematic review tasks.

## Conclusion

Neither LLM achieved the required level of performance for unsupervised use. GPT-4 performed better overall, while Kimi excelled in identifying true positives but required single-article batches for optimal results.

**Tables & Figures**

Table 1

Protocol of Question Items

| Item ID | Question |
| --- | --- |
| 1 | Is technology-based assessment an important topic (i.e., there is enough information about it) of this article? |
| 2 | Is it about learning outcomes? |
| 3 | Is it about learner-centered assessment? |
| 4 | Does it address pre-college education? |
| 5 | Does it specify a geographic context? |
| 6 | Does it offer substantial content about tests measuring learning outcomes? |
| 7 | Does it offer substantial content about tests supporting learning? |
| 8 | Does it offer substantial content about tests supporting teaching? |
| 9 | Does it offer substantial content about tests supporting non-teaching? |
| 10 | Does it offer substantial content about test-associated costs? |
| 11 | Does it offer substantial content about test-associated time? |

*Note*. Items 1-5 correspond to inclusion criteria in the systematic review. Items 6-11 correspond to questions for coding text information in the systematic review.

Table 2

Overview: Performance by LLM

| Metric | Overall | | Averaged from Coder-Specific Statistics | |
|---|---|---|---|---|
| | GPT-4 | Kimi | GPT-4 | Kimi |
| Accuracy | 0.660 | 0.633 | 0.720 (0.029) | 0.690 (0.074) |
| Precision | 0.779 | 0.788 | 0.780 (0.021) | 0.791 (0.034) |
| Recall | 0.791 | 0.712 | 0.791 (0.061) | 0.712 (0.158) |
| F1 Score | 0.785 | 0.748 | 0.784 (0.030) | 0.738 (0.107) |

*Note.* Values in parentheses are standard deviations across coders.

Table 3

Performance by LLM, By Batch Size

| Batch Size | Metric | GPT-4 | Kimi |
|---|---|---|---|
| 1 article | Accuracy | 0.738 | 0.744 |
| | Precision | 0.780 | 0.811 |
| | Recall | 0.827 | 0.787 |
| | F1 Score | 0.803 | 0.799 |
| 4 articles | Accuracy | 0.712 | 0.678 |
| | Precision | 0.775 | 0.783 |
| | Recall | 0.781 | 0.694 |
| | F1 Score | 0.778 | 0.736 |
| 8 articles | Accuracy | 0.711 | 0.648 |
| | Precision | 0.782 | 0.766 |
| | Recall | 0.765 | 0.654 |
| | F1 Score | 0.774 | 0.706 |

Table 4

Inter-coder Reliability

| Batch Size | LLM | ICC | Kappa |
|---|---|---|---|
| 1 article | GPT-4 | 0.815 | 0.468 |
| | Kimi | 0.845 | 0.521 |
| 4 articles | GPT-4 | 0.816 | 0.469 |
| | Kimi | 0.765 | 0.393 |
| 8 articles | GPT-4 | 0.813 | 0.464 |
| | Kimi | 0.649 | 0.269 |

*Note*. ICC = Intra-Class Correlation. Statistics here are based on five sets of codes.

Table 5

Performance by LLM, By Question Item

| Item ID | LLM | Accuracy | Precision | Recall | F1 Score |
|---------|-------|----------|-----------|--------|----------|
| 1 | GPT-4 | 0.964 | 1.000 | 0.964 | 0.981 |
|   | Kimi | 0.836 | 1.000 | 0.836 | 0.911 |
| 2 | GPT-4 | 0.924 | 1.000 | 0.924 | 0.961 |
|   | Kimi | 0.880 | 1.000 | 0.880 | 0.936 |
| 3 | GPT-4 | 0.747 | 1.000 | 0.747 | 0.855 |
|   | Kimi | 0.570 | 1.000 | 0.570 | 0.726 |
| 4 | GPT-4 | 0.747 | 1.000 | 0.747 | 0.855 |
|   | Kimi | 0.714 | 1.000 | 0.714 | 0.833 |
| 5 | GPT-4 | 0.799 | 1.000 | 0.799 | 0.889 |
|   | Kimi | 0.839 | 1.000 | 0.839 | 0.912 |
| 6 | GPT-4 | 0.492 | 0.459 | 0.893 | 0.606 |
|   | Kimi | 0.487 | 0.449 | 0.756 | 0.563 |
| 7 | GPT-4 | 0.492 | 0.459 | 0.905 | 0.609 |
|   | Kimi | 0.526 | 0.476 | 0.810 | 0.599 |
| 8 | GPT-4 | 0.628 | 0.625 | 0.843 | 0.718 |
|   | Kimi | 0.591 | 0.631 | 0.657 | 0.644 |
| 9 | GPT-4 | 0.690 | 0.533 | 0.067 | 0.119 |
|   | Kimi | 0.690 | 0.667 | 0.017 | 0.033 |
| 10 | GPT-4 | 0.810 | 0.367 | 0.300 | 0.330 |
|   | Kimi | 0.841 | 0.478 | 0.183 | 0.265 |
| 11 | GPT-4 | 0.628 | 0.259 | 0.528 | 0.347 |
|   | Kimi | 0.620 | 0.280 | 0.653 | 0.392 |

*Note*. Items 1-5 and Items 6-11 correspond to inclusion criteria and synthesis questions in the systematic review, respectively.

Figure 1

Performance by LLM, By Question Item (Batch Size = 1)

Accuracy

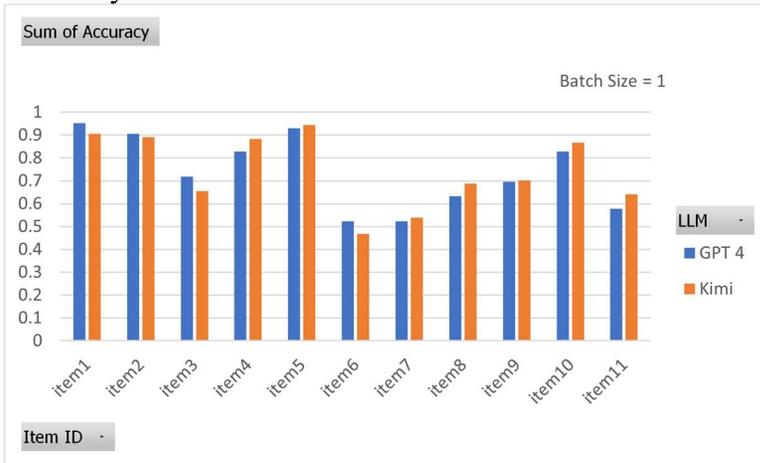

Precision

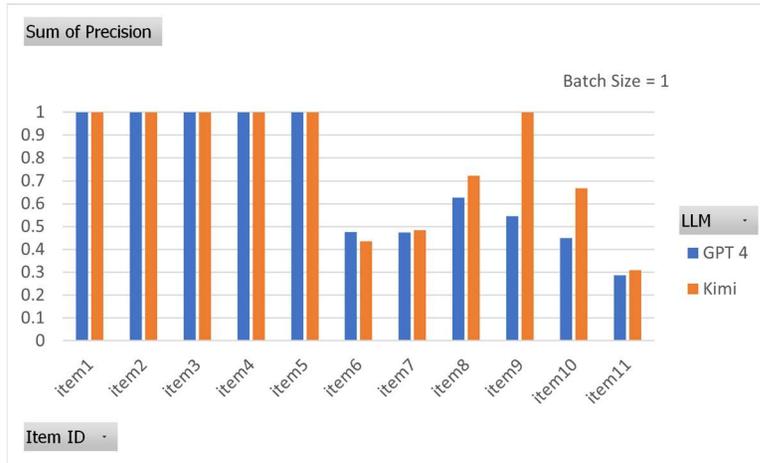

Recall

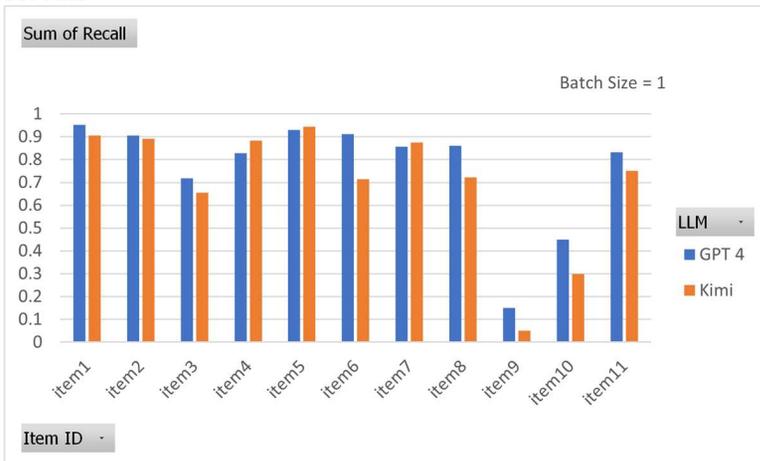

F1 Score

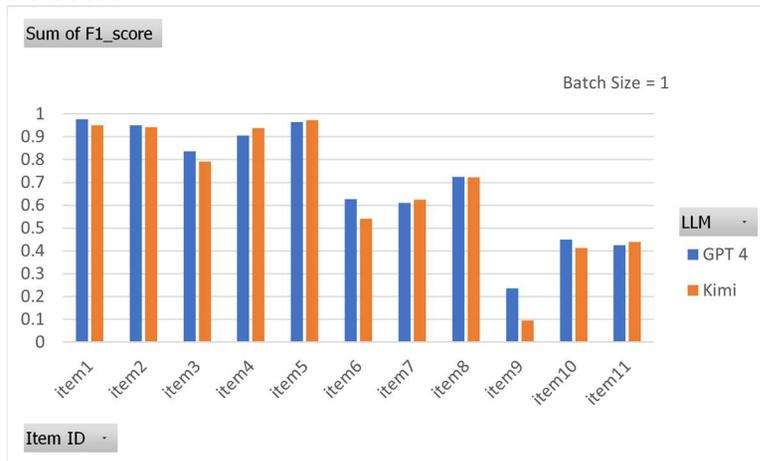

## Appendix A

**Definitions**

In this study, a false positive case occurs when the human-generated code is 0, but the LLM-generated code is 1, indicating that the model identified something that wasn't present. Conversely, a false negative case occurs when the human-generated code is 1 and the LLM-generated code is 0, meaning the model missed relevant information. The term "positive" relates to the nature of the questions we used, which are structured as two-sided, "is it or not" prompts, akin to two-sided hypothesis testing. The null hypothesis posits that "XXX does not have YYY," and when the machine responds "yes," it is treated as rejecting the null hypothesis, signaling a positive outcome.

**Computational Formulae**

Accuracy: Accuracy represents the percentage of correctly classified segments, encompassing both ischemic and non-ischemic segments. It provides a holistic view of the performance, defined as a fraction of correct predictions as

$$\text{Accuracy} = \frac{Number\ of\ correct\ predictions}{Total\ number\ of\ predictions} = \frac{TP+TN}{TP+TN+FP+F} \qquad (1)$$

Precision (Positive Predictive Value): Precision quantifies the accuracy of the positive predictions. It is defined as the ratio of true positive predictions to the total number of positive predictions (including both true positives and false positives). High precision indicates that a high proportion of segments classified as ischemic are indeed ischemic, reducing the likelihood of false alarms, written as

$$Precision = \frac{True\ Positives\ (TP)}{True\ Positives\ (TP)\ + False\ Positives\ (FP)} \qquad (2)$$

Recall: It measures the ability of a model to correctly identify positive cases. It is calculated as the ratio of true positives (correctly predicted positives) to the sum of true positives and false negatives (missed positives). Recall focuses on the model's capacity to detect all positive cases, written as

$$Recall \ = \frac{True \ Positives \ (TP)}{True \ Positives \ (TP) + Fal \quad Negatives \ (FN)} \qquad (3)$$

F1 Score: The F1 score is the harmonic mean of precision and sensitivity, providing a balanced measure of the two. It is particularly useful when considering both false positives and false negatives. A high F1 score indicates a model with both high precision and high sensitivity, ensuring reliable performance in identifying ischemic segments.

$$F1 \ = 2 \times \frac{Precision \times \text{Sensitivity}}{Precision \ + \text{Sensitivity}} \qquad (4)$$